\documentclass[11pt]{article}

\usepackage[letterpaper,textwidth=5.5in,textheight=9in,top=1in]{geometry}
\usepackage[T1]{fontenc}
\usepackage{times}
\usepackage[numbers,sort&compress]{natbib}

\usepackage{amsmath,amssymb}
\usepackage{graphicx}
\usepackage{booktabs}
\usepackage{tabularx}
\usepackage{adjustbox}
\usepackage{xcolor}
\usepackage{url}
\usepackage[hidelinks]{hyperref}
\usepackage{microtype}

\newcommand{\arm}[1]{\textsc{#1}}
\newcommand{\None}{\arm{NoTheory}}
\newcommand{\Oracle}{\arm{Oracle}}
\newcommand{\Irr}{\arm{Irrelevant}}
\newcommand{\Hint}{\arm{GenericHint}}
\newcommand{\Ex}{\arm{Examples}}
\newcommand{\Weak}{\arm{Induced-weak}}
\newcommand{\Strong}{\arm{Induced-strong}}
\newcommand{\Merged}{\arm{Merged}}

\title{What Is a Theory Worth to an LLM Solver?\\Measuring, Inducing, and Merging Theories in Context}

\author{Jun Gao \qquad Shuchao Bi \\[2pt] Meta Platforms, Inc.}

\begin{document}
\maketitle

\begin{abstract}
Mathematicians solve problems faster when they know the right theory. We ask how much a stated theory is worth to a large language model (LLM) solver, and whether a model can write that theory itself. In a length-matched protocol, adding the right theory to the context of Qwen3-14B raises pass@1 on pattern waiting-time problems from 0.37 to 0.83 and cuts the median tokens of correct solutions from 10.2k to 6.3k, while worked examples of the same length do not help. The gain shrinks as the solver and its token budget grow (+0.46, +0.26, +0.07), and at a long budget it comes almost entirely from finishing within budget: solutions that finish are correct 98--100\% of the time in every condition. To separate induction from recall, we plant unpublished theorems in custom probability families and measure how much of the oracle theory's gain an induced theory recovers. A strong open model, required to explore the training problems with code before writing, recovers 60\% of the gain on one planted family, 48\% on a second, and almost none on a third, where it finds the right form but not the condition that makes the form computable. No single induced note is complete, but a strong model that merges several notes recovers 61\% of the gain on the classic family, above the best single note (47\%). Theories are worth the reasoning they save, and models can find their pieces before they can assemble them.
\end{abstract}

\section{Introduction}
\label{sec:intro}

A student who knows Conway's leading numbers computes the expected waiting time for a coin pattern in a few lines; a student who does not sets up and solves a linear system over the pattern's prefixes. Both can be right, but only one is efficient. Large language models (LLMs) increasingly solve such problems with long chains of thought~\cite{wei2022cot}, and their resource use grows with the reasoning they must produce~\cite{snell2025scaling}. This raises two questions. First, how much is a stated theory worth to an LLM solver: how much accuracy does it add and how much reasoning does it save? Second, can a model induce such a theory from problems alone, so that it uses the compute of discovery once rather than on every problem?

Answering either question cleanly is harder than it looks. Adding text to the context changes its length; examples and theory can carry the same information; and a model that ``induces'' a classic theorem may simply be recalling it. We address these confounds with three design choices. (i)~All conditions use contexts of the same length, and we compare the oracle theory with irrelevant text, a generic hint, and worked examples of the same length. (ii)~We report the \emph{recovery ratio} $R$, the share of the oracle theory's accuracy gain that an induced theory recovers. (iii)~To separate induction from recall, we \emph{plant} unpublished theorems in custom probability families, verify that no model can state them before seeing the problems, and only then let models induce theories from training problems.

\paragraph{Contributions.}
\begin{itemize}
    \item A controlled protocol for measuring the value of a theory in context, with length-matched controls, paired bootstrap intervals and a recovery ratio (Section~\ref{sec:setup}).
    \item Evidence that a stated theory makes a solver both more accurate and cheaper, that worked examples do not substitute for it, and that at long budgets its value is almost entirely the reasoning it saves (Sections~\ref{sec:h1} and~\ref{sec:mechanism}).
    \item \textbf{A planted-theorem design that separates induction from recall:} custom probability families with unpublished theorems that no model can state before seeing the problems, with evidence that a strong open model can induce such theorems when forced to explore with code, and a characterization of how it fails: the right form without the condition, or a special case without the general law (Section~\ref{sec:h2}).
    \item Evidence that no single induced theory is complete but that a strong model can merge partial theories from several inducers, without test information, into one that passes our success bar (Section~\ref{sec:merge}).
\end{itemize}

\section{Setup}
\label{sec:setup}

\subsection{Problem families}
Every family has an automatically graded numeric answer, a key theory that makes the problems easy, a generator for fresh parameters, and a solver accuracy without the theory that leaves room for improvement. Table~\ref{tab:families} summarizes them.

\begin{itemize}
    \item \textbf{Family (a): pattern waiting times (classic).} Expected waiting times for coin patterns, probabilities that one pattern appears before another, and waiting times for either of two patterns. The key theory is Conway's overlap numbers and the martingale method~\cite{li1980martingale,guibas1981string}. Every answer is computed twice, from the overlap numbers and from the pattern automaton, and the two must agree. We use 40 training and 160 test problems at the difficulty where the solver without theory scores 0.38 in calibration.
    \item \textbf{Family (b): symmetric colouring counts (classic).} Counting colourings up to symmetry, with Burnside's lemma and P\'olya enumeration~\cite{polya1937kombinatorische} as the theory; 156 problems, about 20\% of them traps.
    \item \textbf{F1: carry-over rearrangements on a necklace (planted).} A ring of $L$ sites with $K$ colours. Each window of three sites $(x,y,z)$ moves forward to $(y,z,x\oplus1)$ at rate $f(x,y,z)$ and backward at rate $g$, so the chain is not reversible. The planted theorem states that if a weighted flux is constant on every orbit of the window map, the stationary law is a product law with explicit weights. Held-out problems have between $1.4\times10^7$ and $3.9\times10^8$ states.
    \item \textbf{F2: carry-over swaps of neighbouring pairs (planted).} A pair $(x,y)$ becomes $(y,x\oplus1)$ at a rate that also depends on the two outer neighbours. The planted theorem states that if $f(u,x,y,v)\,M(u,x)M(x,y)M(y,v)$ is constant on every orbit of the pair map, the stationary law is a nearest-neighbour (transfer-matrix) law $\pi(\eta)\propto\prod_j M(\eta_j,\eta_{j+1})$, not a product law.
    \item \textbf{F3: a random collision process (planted).} Pairs of particles of $K$ kinds collide and either vanish or turn into another pair. The planted theorem states that if weights $b$ satisfy one linear equation per pair of kinds, the expected total penalty until the box is empty is $\sum_c b_c n_c$, by optional stopping on a linear potential.
\end{itemize}

For the planted families, the exact solution equals the planted theorem on every verification instance (F2 48/48, F3 48/48), about 20\% of held-out problems are traps on which the theorem does not apply, and misapplying the theorem on a trap gives one predictable wrong number, so misapplication is measurable. A planted family is used only if the oracle theory raises held-out accuracy by at least 0.20 in calibration (F2 +0.83, F3 +0.58). As a classic control under the identical planted-family protocol, we also run family (a) (``P5b'').

\begin{table}[t]
\centering
\caption{Problem families. ``Oracle gain'' is held-out pass@1 with the oracle theory minus without it, for the confirmatory solver and cap (Table~\ref{tab:main}).}
\label{tab:families}
\small
\begin{tabularx}{\textwidth}{@{}l l X l@{}}
\toprule
\textbf{Family} & \textbf{Kind} & \textbf{Theory} & \textbf{Oracle gain} \\
\midrule
(a) waiting times & classic & Conway overlap numbers / martingale & +0.46 (12k) \\
(b) colouring counts & classic & Burnside / P\'olya & solver already knows it \\
F1 necklace rearrangements & planted & orbit-flux condition $\Rightarrow$ product law & +0.76 (12k) \\
F2 neighbouring-pair swaps & planted & orbit condition $\Rightarrow$ transfer-matrix law & +0.82 (32k) \\
F3 collision process & planted & linear potential $\Rightarrow$ expected penalty & +0.63 (32k) \\
P5b = (a) under planted protocol & classic & Conway overlap numbers & +0.35 (32k) \\
\bottomrule
\end{tabularx}
\end{table}

\subsection{Conditions}
The solver is fixed within each comparison and sees one of the following contexts, all padded with neutral text to the same token count: \None{} (padding only); \Oracle{} (the human-written theory; in family (a) it includes one worked example that is not a test problem); \Irr{} (an unrelated theory of the same length; family (a)); \Hint{} (a generic hint to look for a product form or invariant; planted families); \Ex{} (worked training solutions that use the method; family (a)); \Weak{} and \Strong{} (theories induced by Qwen3-32B and by Qwen3-235B-A22B-Thinking-2507~\cite{yang2025qwen3}, respectively; in family (a), the strong inducers also include the original Qwen3-235B-A22B); and \Merged{} (a theory produced by a model that merges several induced theories). Appendix~\ref{app:arms} maps these names to the condition labels of our preregistrations.

\paragraph{Induction.} An inducer sees only the training problems and its own solution attempts, may run code on small instances, and writes a theory no longer than the oracle text that states when it applies and contains no answer to a specific problem. It never sees the planted theorem or any held-out problem. We run three independent inductions per inducer. In our first protocol, models wrote theories without running any code, and their theories were guesses; even the strong model attempted to write a theory before any exploration in all three runs. Our second protocol, used for all planted-family results, shows the model how to read the data and accepts a theory only after at least three code runs. In family (a), induction instead generates several candidate theories and selects among them on a validation split of the training problems (Appendix~\ref{app:prompts}).

\paragraph{Solvers and sampling.} Qwen3-14B in thinking mode (temperature 0.6, top-$p$ 0.95, top-$k$ 20) is the main solver, at token caps of 12k and 32k; Qwen3.6-27B at 32k and 64k is a second, stronger solver. We draw $k=8$ samples per problem in family (a) and $k=3$ in the planted families.

\subsection{Metrics}
We report pass@1, the median number of tokens of correct solutions, and the share of samples truncated at the cap. The recovery ratio of an induced theory is
\begin{equation}
R \;=\; \frac{\mathrm{acc}(\text{induced}) - \mathrm{acc}(\None)}{\mathrm{acc}(\Oracle) - \mathrm{acc}(\None)},
\end{equation}
computed on non-trap held-out problems with 95\% paired bootstrap intervals; with three inductions, $R$ uses the mean accuracy over the three theories. Induction succeeds (H2) if $R \geq 0.5$ and the induced theory beats \Hint{} (or \Irr{}) with an interval excluding zero. On traps we report the misapplication rate. A blind LLM judge classifies each induced theory against the planted theorem as (i) the same, (ii) equivalent, (iii) a partial or special case, (iv) wrong but useful, or (v) wrong, by two-of-three majority. Two judges agree poorly (Appendix~\ref{app:judge}), so judge classes are descriptive and solver accuracy is the primary evidence.

\paragraph{Recall controls.} Before any inducer sees the problems, we ask it whether it knows a closed form for the family (the contamination probe). In family (a) we also rename symbols and stories. For the planted families, no probe answer states the planted condition (Section~\ref{sec:h2}).

\section{A theory in context makes the solver more accurate and cheaper}
\label{sec:h1}

\begin{figure}[t]
  \centering
  \includegraphics[width=\textwidth]{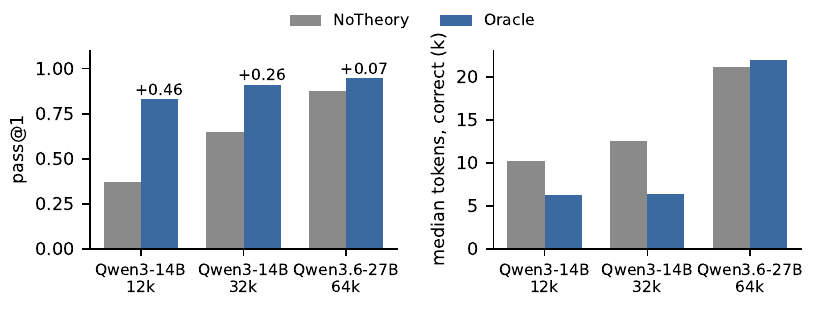}
  \caption{Family (a): pass@1 (left) and median tokens of correct solutions (right) without and with the oracle theory, for three solver and cap settings. The accuracy gain shrinks as the solver and budget grow. The theory halves tokens at 12k and 32k and matches \None{} at 64k (where it is about 30\% cheaper than \Irr{} and a searched induced theory).}
  \label{fig:h1}
\end{figure}

On family (a) with Qwen3-14B at a 12k cap, the oracle theory raises pass@1 from 0.37 to 0.83 (paired difference $+0.46$ [$+0.41$, $+0.52$]) and pass@8 from 0.68 to 0.99. Correct solutions use a median of 6.3k tokens instead of 10.2k, and the share of samples truncated at the cap falls from 59\% to 14\% (Figure~\ref{fig:h1}). Length alone does nothing: an irrelevant theory of the same length changes pass@1 by $-0.02$ (interval including zero). Worked examples do not substitute for the theory either: training solutions that use exactly the method change pass@1 by $-0.02$, and the theory beats them by $+0.48$. The gain is not surface matching: with renamed symbols and stories it is $+0.51$, against $+0.47$ on the same original problems.

\paragraph{The gain shrinks as the solver gets stronger.} The 12k cap truncates most solutions without the theory, so part of the gain is finishing within budget. With a 32k cap the same solver without the theory reaches 0.65 and truncation falls to 1\%; the theory still adds $+0.26$ [$+0.22$, $+0.31$] and halves the tokens of correct solutions (12.6k to 6.4k). Qwen3.6-27B at a 64k cap is close to the ceiling without the theory (0.88); the theory adds $+0.07$ [$+0.04$, $+0.10$], and its correct solutions are about as long as those without it (22.0k vs.\ 21.1k) but about 30\% shorter than with the irrelevant theory or a searched induced theory (29--33k).

\paragraph{Negative control: a theory the solver already knows.} On family (b), Qwen3-14B scores 0.86--0.95 without the theory, and even with Qwen3-4B the Burnside/P\'olya theory adds only $+0.022$ (not significant) at 12k and 32k; trap problems did not induce the solver to misapply it. A theory is worth what the solver does not already know.

\section{Can a model induce an unpublished theory?}
\label{sec:h2}

\begin{table}[t]
\centering
\caption{Induced theories across families (non-trap held-out accuracy, $k=3$). The three numbers in the induced columns are the three independent inductions. $R$ is reported with 95\% paired bootstrap intervals; H2 requires $R_{\Strong}\geq0.5$ and $\Strong>\Hint$. Judge classes are the blind classification of the three \Strong{} theories by the primary judge (gpt-oss-120b) and, after the slash, the secondary judge (Muse Spark 1.3); the two agree poorly (Appendix~\ref{app:judge}). Generated from the result files by a script (Appendix~\ref{app:repro}).}
\label{tab:main}
\scriptsize
\begin{adjustbox}{max width=\textwidth}
\begin{tabular}{@{}llccccccl@{}}
\toprule
\textbf{Family} & \textbf{Solver, cap} & \None & \Hint & \Oracle & \Weak & \Strong & $R_{\Weak}$ / $R_{\Strong}$ & \textbf{H2; judge} \\
\midrule
F1 & Qwen3-14B, 12k & 0.02 & 0.08 & 0.78 & .17/.69/.10 & .40/.43/.59 & 0.39 [.34,.46] / 0.60 [.51,.68] & met; ii, split, iii / -- \\
F1 & Qwen3-14B, 32k & 0.16 & 0.28 & 0.94 & .40/.87/.26 & .62/.70/.66 & 0.45 [.37,.53] / 0.65 [.55,.74] & (robustness) \\
F1 & Qwen3.6-27B, 32k & 0.01 & 0.08 & 0.48 & .07/.41/.06 & .15/.04/.31 & 0.35 [.27,.44] / 0.33 [.24,.43] & (robustness) \\
F1 & Qwen3.6-27B, 64k & 0.58 & 0.65 & 0.99 & -- & .88/.69/.65 & -- / 0.40 [.24,.54] & (mechanism) \\
F2 & Qwen3-14B, 32k & 0.00 & 0.01 & 0.82 & .00/.00/.00 & .00/.00/.01 & 0.00 [.00,.00] / 0.00 [.00,.01] & fails; v, iii, iii / iii, iii, v \\
F3 & Qwen3-14B, 32k & 0.37 & 0.72 & 1.00 & .15/.65/.18 & .01/1.00/.99 & $-$0.07 [$-$.24,.07] / 0.48 [.38,.54] & not met; v, ii, ii / v, ii, i \\
P5b & Qwen3-14B, 32k & 0.56 & 0.60 & 0.91 & .44/.55/.58 & .74/.96/.82 & $-$0.11 [$-$.41,.10] / 0.81 [.67,.96] & met; iii, iii, iii / iii, ii, iii \\
\bottomrule
\end{tabular}
\end{adjustbox}
\end{table}

\begin{figure}[t]
  \centering
  \includegraphics[width=\textwidth]{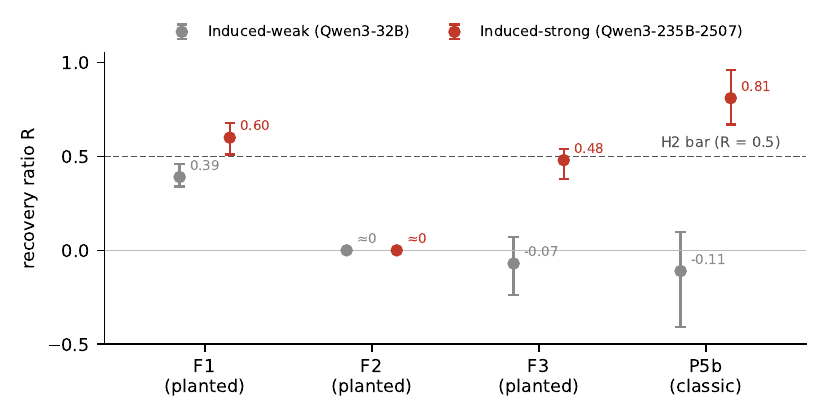}
  \caption{Recovery ratio $R$ by family and inducer (confirmatory settings of Table~\ref{tab:main}), with 95\% intervals. The strong inducer passes the bar on one planted family (F1) and on the classic control (P5b), reaches it on average on F3 with one harmful theory, and fails on F2.}
  \label{fig:r}
\end{figure}

Table~\ref{tab:main} and Figure~\ref{fig:r} give the results across families. The strong inducer succeeds on F1, comes close on F3, fails on F2, and succeeds on the classic control, where recall cannot be ruled out. The weak inducer recovers 39\% on F1 and essentially nothing elsewhere.

\paragraph{F1: induction succeeds.} The strong inducer's theories recover $R=0.60$ [0.51, 0.68] of the oracle gain and beat the generic hint by $+0.40$ [0.32, 0.48]. Two of its three runs rediscover the planted condition: executed literally, their rules agree with the planted theorem on all 30 training problems, all 48 held-out non-trap problems, and all 12 traps, which they correctly reject. The best single theory comes from the weak inducer (the planted condition in reciprocal form, $R=0.89$), but the weak inducer finds the rule in one run of three and the strong inducer in two. The result holds at a 32k cap ($R=0.65$). With Qwen3.6-27B at 32k it falls to 0.33, because 99\% of that solver's samples without the theory and 53\% with it are truncated; of the solutions that finish, 284 of 285 are correct. A post-hoc analysis at 12k shows that the largest loss in every condition is failing to write an answer within the budget; one strong theory's recipe is off by one in a single step, and exactly 18 answers equal the resulting wrong value; and problems with backward moves are hard for every condition, including the oracle (29 of 54 correct).

\paragraph{This is induction, not recall.} Before seeing any problem, neither inducer states the planted condition: all 16 probe answers reason through detailed balance, and none gives the condition. The models' prior is that the beads are independent; the condition is induced from the data.

\paragraph{F2: the right form without the condition.} On F2 the strong inducer recovers $R=0.003$ [0.000, 0.008]. Two of its three theories state the correct nearest-neighbour form, but neither states the orbit condition that determines the weight matrix $M$, so the solver cannot compute $M$ and the theory is useless (Box~\ref{box:f2}). Other inducers guess the wrong class: all three weak theories and all three theories of the original Qwen3-235B-A22B assert that the beads are independent. In the probes, no strong-model answer gives the pair form.

\paragraph{F3: a special case, and a harmful theory.} On F3 the strong inducer recovers $R=0.48$ [0.38, 0.54] but does not beat the generic hint ($-0.05$ [$-0.13$, 0.05]): the hint to look for an invariant already lifts accuracy from 0.37 to 0.72. Two of its three theories state the planted theorem completely (accuracies 1.00 and 0.99, as high as the oracle), but the third states a wrong product formula and drives the solver below no theory (0.01). A weak theory that states only the step-count special case helps on questions about the number of collisions (0.99) but hurts on questions about pair-dependent penalties (0.32, against 0.49 without theory). None of the 16 probe answers (8 strong, 8 weak, with the wording corrected to name F3's own quantity) gives the general penalty law: all eight strong answers and five weak ones give only the special case of equal weights, and one weak answer gives the step-count special case, which follows from textbook optional stopping.

\paragraph{The classic control.} On P5b the strong inducer recovers $R=0.81$ [0.67, 0.96] and beats the hint by $+0.24$, but two of four probe answers already name Conway,\footnote{This strong probe used the stationary-distribution wording of the planted families, but its answers still discussed waiting times. With the corrected wording, which names the waiting times, none of the 8 weak-inducer answers names Conway or gives the race formula; the strong probe was not rerun.} so this success cannot be attributed to induction. The contrast with the planted families estimates how much recall helps: a classic theory that the model may have seen is recovered better than any planted one.

\begin{figure}[t]
\fbox{\begin{minipage}{0.97\textwidth}\small
\textbf{Box 1: the right form without the condition (F2, strong inducer, run 1; excerpt).}\label{box:f2}
``The stationary distribution has a \emph{Markov chain product form}: $\pi(\sigma)\propto\prod_{i=1}^{n} M(\sigma_{i-1},\sigma_i)$, where the transition matrix $M$ satisfies $M(a,b)\propto r^{c(a,b)}$ for a conserved quantity $c(a,b)$ \ldots{} Detailed balance implies $r=s$, defining the balance ratio $r$.''

\smallskip
The form is the planted transfer-matrix law, but $M$ is tied to a ``balance ratio'' from detailed balance rather than to the orbit condition $f(u,x,y,v)M(u,x)M(x,y)M(y,v)=\text{const}$ on orbits, so the solver cannot recover $M$ (held-out accuracy 0.00).
\end{minipage}}
\end{figure}

\paragraph{Family (a): searching helps a weaker inducer, but it misses the key rule.} On the classic family, a single theory induced by Qwen3-32B is harmful (pass@1 $-0.11$ relative to \None{} at 12k): it restates a generic state-machine method and suppresses the solver's own shortcuts. Generating several candidate theories and selecting among them on a validation split of the training problems turns the same model's theory into a useful one ($R=0.28$ [0.18, 0.37] at 12k; a refined candidate reaches 0.47 [0.34, 0.59] at 32k). Validation accuracy predicts test accuracy across candidates (Spearman 0.93), so selection on training data is reliable. But no 32B candidate finds the cross-overlap rule between two patterns, which is the heart of Conway's method; stronger inducers alone recover 0.32--0.34 at 32k.

\begin{table}[t]
\centering
\caption{How induced theories fail, with examples from our runs.}
\label{tab:failures}
\small
\begin{tabularx}{\textwidth}{@{}l X X@{}}
\toprule
\textbf{Failure mode} & \textbf{Example} & \textbf{Effect on the solver} \\
\midrule
No exploration & First protocol: theories written without running code & Guesses; no planted condition found \\
Wrong class & F2: weak and original-235B theories assert independent beads & $R \approx 0$ \\
Right form, missing condition & F2: transfer-matrix form without the orbit condition (Box~1) & $R=0.003$: the solver cannot compute $M$ \\
Special case only & F3 weak run 1: step count, not pair-dependent penalty & Helps on step counts (0.99), hurts on penalties (0.32 vs.\ 0.49) \\
Confident wrong formula & F3 strong run 0: wrong product formula & Below no theory (0.01 vs.\ 0.37) \\
Partial sub-rules & Family (a): each inducer finds a different sub-rule & Mergeable (Section~\ref{sec:merge}) \\
\bottomrule
\end{tabularx}
\end{table}

Table~\ref{tab:failures} summarizes the failure modes we observe. The most instructive is the second F2 failure: a theory can have the right mathematical form and still be useless, because a solver needs the condition that pins down the form's parameters. Accuracy downstream, not resemblance to the true theorem, is the test of a theory.

\paragraph{Exploration is necessary.} Under our first protocol, inducers wrote theories without running any code, and the theories were guesses. Under the second, which accepts a theory only after three code runs, the same strong model induced the F1 condition. Requiring evidence of exploration before writing was the single most important protocol change.

\section{Merging partial theories}
\label{sec:merge}

\begin{figure}[t]
  \centering
  \includegraphics[width=0.62\textwidth]{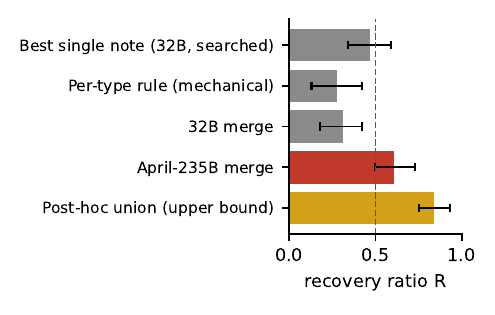}
  \caption{Family (a) at 32k: no single induced note passes the bar, but a strong model that merges 12 notes (with training problems only) does. The post-hoc union of the correct sub-rules, chosen by hand, is an upper bound.}
  \label{fig:merge}
\end{figure}

On family (a), no inducer writes a complete theory. Different inducers find different sub-rules: Qwen3-235B-2507 finds the rule for one pattern before another, and the original Qwen3-235B-A22B, given a longer budget, finds the rule for either of two patterns. At a 32k cap, the best single induced note recovers $R=0.47$ [0.34, 0.59]. A hand-made union of the correct sub-rules from different inducers recovers 0.84 [0.75, 0.93], which shows that the pieces exist but are spread across notes.

Can a model do the assembly? We give a merger 12 induced notes and the training problems, with no test information, and ask for one theory of the same length. The original Qwen3-235B-A22B merges them into a theory with $R=0.61$ [0.50, 0.73] (three merges: 0.70, 0.57, 0.56), which beats the irrelevant theory by $+0.19$ and passes our success bar. Qwen3-32B as merger reaches 0.31, and a mechanical rule that picks the best note for each problem type reaches 0.28 (Figure~\ref{fig:merge}). Merging is therefore a reasoning task in its own right: a strong merger reconciles partial theories better than either a weak merger or a selection rule.

\section{Mechanism: theories save reasoning, not correctness}
\label{sec:mechanism}

\begin{figure}[t]
  \centering
  \includegraphics[width=0.48\textwidth]{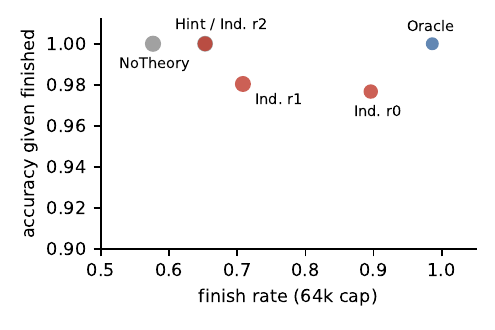}
  \caption{F1 with Qwen3.6-27B at a 64k cap. Every condition is correct on 98--100\% of the solutions that finish; conditions differ in how often the solver finishes. Marker size is proportional to tokens per sample (35k with the oracle theory, 54k without). Ind.\ = \Strong{} (three runs); Hint = \Hint{}.}
  \label{fig:mechanism}
\end{figure}

If a theory only helps the solver finish, a large enough budget should remove its accuracy advantage but not its compute advantage. We test this on F1 with Qwen3.6-27B at a 64k cap (Figure~\ref{fig:mechanism}). Without the theory the solver finishes 58\% of samples, and every finished sample is correct; with the oracle theory it finishes 99\%, using 35k tokens per sample instead of 54k. The strong inducer's theories finish 65--90\% of samples and are correct on 98--100\% of those that finish, so the induced theories are applied correctly almost every time; what separates them from the oracle is how much reasoning they save (43--53k tokens). The same pattern holds on family (a) at 64k, where finished solutions are correct 92\% of the time without the theory and 98--99\% with any context.

\paragraph{Traps.} A theory can also be misapplied. On trap problems, the solver with the oracle theory misapplies it on 22\% of F2 traps and 47\% of F3 traps; on F2, the other conditions almost never do; on F1, every condition scores near zero on traps, which are therefore uninformative there.

\paragraph{Judging.} Judge classes are weaker evidence than solver accuracy. Given identical anonymized inputs, the preregistered primary judge (gpt-oss-120b) and a secondary judge (Muse Spark 1.3) agree on only 10 of the 21 induced theories of F2, F3 and P5b (Cohen's $\kappa=0.11$; $0.18$ after merging the classes into recovered, partial and wrong). On F3 both judges track accuracy (Spearman $\rho=0.95$ and $0.93$): they rate the two strong theories with oracle-level accuracy as the same as or equivalent to the planted theorem, and the harmful one as wrong. On P5b they disagree. The secondary judge rates as equivalent only the theory that states the closed form for either pattern, in line with accuracy ($\rho=0.66$); the primary judge rates all three strong theories as partial and two weak theories as equivalent, against accuracy ($\rho=-0.62$). On F2 accuracy is at the floor, so tracking cannot be tested. We therefore treat judge classes as descriptive and base every conclusion on solver accuracy (Appendix~\ref{app:judge}).

\section{Discussion}
\label{sec:discussion}

\paragraph{The value of a theory is the reasoning it saves.} Across solvers and budgets, the accuracy gain from a theory shrinks, but it does so because the solver without the theory can eventually reason its way to the same answer. At a long budget, solutions that finish are almost always correct with or without a theory, and what the theory changes is how often the solver finishes and how many tokens it spends. For deployed systems, where budgets are fixed and tokens are paid for, this means a theory's value should be measured in both accuracy and compute, and the compute side persists when the accuracy side vanishes.

\paragraph{Discovery can be paid once.} Inducing a theory uses one exploration on training problems; using it saves reasoning on every subsequent problem. Our results show that this amortization works when the inducer explores the data (F1) and that it can fail silently when the theory lacks the condition that makes it computable (F2). A practical pipeline should therefore validate an induced theory by its effect on a solver, as our recovery ratio does, rather than by how plausible it reads.

\paragraph{Pieces before wholes.} Induced theories are often partial, and different inducers find different parts. Merging by a strong model recovers more than any single theory, while merging by a weak model or by a selection rule does not. This suggests treating theory induction as a population process: induce several partial theories, then compose them with a strong model, and select on held-out training problems.

\section{Related work}
\label{sec:related}

\paragraph{Knowledge and guidance in context.} Retrieval augmentation adds external knowledge to the context of a language model~\cite{lewis2020rag}. For reasoning, models benefit from step-by-step solutions~\cite{wei2022cot}, from abstracting a problem into its governing principle before solving it~\cite{zheng2024stepback}, from self-generated related examples~\cite{yasunaga2024analogical}, and from libraries of reusable thought templates~\cite{yang2024bot}. We isolate the value of one ingredient, a stated theory, with length-matched controls, and we find that worked examples using the same method do not substitute for it.

\paragraph{Test-time compute.} Spending more reasoning at test time can be more effective than scaling model parameters~\cite{snell2025scaling}. Our results describe the complementary resource: a stated theory reduces the reasoning a problem requires, and at long budgets this reduction is most of its value.

\paragraph{Induction and discovery with language models.} Language models can propose and refine hypotheses about input--output rules~\cite{wang2024hypothesis,qiu2024phenomenal} and learn rule libraries~\cite{zhu2023rules}; program search with language models has produced new mathematical constructions~\cite{romeraparedes2024funsearch}; and library learning grows reusable abstractions for program synthesis and theorem proving~\cite{ellis2021dreamcoder,wang2024legoprover}. We study the induction of a mathematical theory in natural language from solved problems, and we measure its value by what it does for a separate solver.

\paragraph{Contamination.} Benchmark results can be inflated by training-data contamination, which can be detected by guided prompting~\cite{golchin2024timetravel} or by statistical tests~\cite{oren2024proving}. We avoid the issue for induction by construction: the planted theorems do not exist in any literature, and probes confirm that models cannot state them beforehand.

\section{Limitations}
\label{sec:limits}
Our families are few and synthetic, and our solvers and inducers are open models from one family; results may differ for other model families and for theories outside probability and combinatorics. The planted families are novel by construction, but their proof techniques (local balance, optional stopping) are textbook, so we cannot exclude partial recall of the method. The two blind judges agree poorly, so judge classes are descriptive only. Eight of the 160 test problems of family (a) are twins of training problems up to swapping heads and tails; excluding them moves every $R$ by at most 0.02. Each condition uses three inductions, so per-run variation is large (Table~\ref{tab:main}).

\bibliographystyle{plainnat}
\bibliography{refs}

\appendix

\section{Condition names}
\label{app:arms}
The paper uses descriptive condition names. The table maps them to the condition identifiers used in the preregistrations and result files of each task, so that every number can be traced to its source.

\begin{table}[h]
\centering
\small
\begin{tabular}{@{}lll@{}}
\toprule
\textbf{Paper name} & \textbf{Task 60 (families (a), (b))} & \textbf{Tasks 61/62 (F1, F2, F3, P5b)} \\
\midrule
\None & C0 & C0 \\
\Oracle & C1 & C1 \\
\Irr & C2 (Burnside/P\'olya text in family (a)) & -- \\
\Hint & -- & C2 \\
\Ex & C3 (worked training solutions) & -- \\
\Weak & C4 (v1), C4a--C4c (searched, v2) & C3 (three runs) \\
\Strong & C4d (Qwen3-235B-2507, original Qwen3-235B) & C4 (three runs); C4a = original Qwen3-235B (F2) \\
\Merged & A10 merges (M\_april, M\_32b, S\_pertype) & -- \\
\bottomrule
\end{tabular}
\end{table}

\section{Preregistration and amendments}
\label{app:prereg}
All analyses were preregistered before the corresponding runs. \textbf{H1} holds if the oracle theory beats both \None{} and the length-matched control on at least two of three metrics and is not worse than worked examples. \textbf{H2} holds if an induced theory recovers at least half of the oracle gain ($R\geq0.5$) and beats the generic hint or irrelevant theory with an interval excluding zero. Follow-up experiments were preregistered as amendments; three outcomes of the amendment process matter for this paper. (i)~The first round of family (b) was voided because colour-swapped duplicates of training problems appeared in the test set; it was regenerated without them. (ii)~A classic control based on commute times in electrical networks was voided by its preregistered calibration rule: the oracle theory did not raise accuracy (0.117 with the theory, 0.133 without), because the solver reported effective resistances without simplifying the network. It was replaced by P5b. (iii)~The 64k run on F1 was registered as a mechanism follow-up, not as a confirmatory test. Table~\ref{tab:prereg} lists the preregistrations, their decision rules and outcomes; the full texts are available from the authors on request.

\begin{table}[h]
\centering
\caption{Preregistrations. Every document was written before any model output of the runs it governs; amendments were dated and written before the outputs they affect.}
\label{tab:prereg}
\scriptsize
\begin{tabularx}{\textwidth}{@{}l X X@{}}
\toprule
\textbf{Study} & \textbf{Decision rule} & \textbf{Amendments and outcome} \\
\midrule
Family (a), H1 and H2 & H1: \Oracle{} beats \None{} and \Irr{} on two of three metrics and is not worse than \Ex{}. H2: $R\geq0.5$ and the induced arm beats \Irr{} (CI excluding 0). & H1 met at 12k, 32k and 64k. A single 32B theory is harmful; searched induction (8 candidates, validation selection) gives $R=0.28$. Stronger inducers: $R\leq0.47$ for any single note at 32k. \\
Family (a), merging & Merged arm: $R\geq0.5$ and beats \Irr{}; per-type selection rule and its keyword patterns fixed in advance. & Met by the 235B merge ($R=0.61$); 32B merge 0.31; per-type rule 0.28. \\
Family (b) & As family (a); about 20\% traps. & First round VOID (colour-swapped train/test twins), regenerated. No gain: the solver already knows the theory. \\
F1 & H2: $R_{\Strong}\geq0.5$ and $\Strong>\Hint$ (CI excluding 0); fails if the upper end of $R$'s CI is below 0.5. VOID if headroom $<0.10$, leak, length mismatch $>1\%$, mixed compute lineages, or fewer than two theories. & Amendment 1: induction protocol v2 (code gate) after the first weak run wrote all theories without running code. H2 met ($R=0.60$). \\
F2, F3 & Calibration gate: \Oracle{} $-$ \None{} $\geq0.20$ on a separate 20-problem bank, else simplify once. H2 and VOID rules as F1. & Both gates passed (+0.83, +0.58). F3 probe corrected to ask about F3's own quantity. F2: H2 fails ($R=0.003$). F3: not met ($R=0.48$, no gain over \Hint{}). \\
Classic control & Commute times in electrical networks, same gate. & Gate failed twice (second attempt: 0.12 with the oracle theory, 0.13 without); stopped as preregistered. Replaced by P5b: H2 met ($R=0.81$), recall not excluded. \\
\bottomrule
\end{tabularx}
\end{table}

\section{Family definitions and planted theorems}
\label{app:families}
\textbf{F1.} A ring of $L\geq3$ sites with colours $0,\ldots,K-1$. Window $i$ is the triple $(\eta_i,\eta_{i+1},\eta_{i+2})=(x,y,z)$; it moves forward to $\tau(x,y,z)=(y,z,x\oplus1)$ at rate $f(x,y,z)$ and backward to $\tau^{-1}$ at rate $g$. Every forward move raises the colour sum by one modulo $K$, so a forward-only chain is not reversible. \emph{Planted theorem:} with $W(x,y,z)=p_xp_yp_z$, if the orbit flux $W(t)f(t)-W(\tau t)g(\tau t)$ is constant on every $\tau$-orbit of triples, the product law $\pi(\eta)\propto\prod_j p_{\eta_j}$ is stationary, and unique if the chain is irreducible.

\textbf{F2.} A ring of $n\geq4$ beads; the pair $(x,y)$ with outer neighbours $(u,v)$ becomes $\sigma(x,y)=(y,x\oplus1)$ at rate $f(u,x,y,v)$. \emph{Planted theorem:} for a positive $K\times K$ matrix $M$, if $f(u,x,y,v)M(u,x)M(x,y)M(y,v)$ is constant on every $\sigma$-orbit of pairs for each $(u,v)$, then $\pi(\eta)\propto\prod_j M(\eta_j,\eta_{j+1})$ is stationary; it is a product law only if $M$ has rank one, which is F1's case. The proof balances the flow into and out of each configuration through each pair window.

\textbf{F3.} A box with an even number of particles of $K$ kinds; at each step two particles are chosen uniformly at random, and a pair of kinds $(c,d)$ either vanishes or becomes another pair, at a penalty $r(c,d)$. \emph{Planted theorem:} if weights $b$ satisfy $b_c+b_d-\mathbb{E}[b(\text{outcome})]=r(c,d)$ for every pair of kinds, the expected total penalty until the box is empty is $\sum_c b_c n_c$; the proof applies optional stopping to the martingale $\Phi_t+C_t$, where $\Phi_t=\sum_c b_cn_c(t)$ and $C_t$ is the penalty accumulated so far. On traps, probability mass on one pair moves from a product to vanishing, so no weight vector satisfies every equation.

\textbf{Verification.} Every answer is computed exactly and checked independently of the theory (Table~\ref{tab:verify}). Data splits use fixed seeds: F1 has 30 training and 60 held-out problems (48 non-traps, 12 traps); F2 30 and 62 (50 non-traps); F3 30 and 60; family (a) 40 and 160. Calibration banks are disjoint from all three and are never shown to an inducer. Sample problem statements are available from the authors on request.

\begin{table}[h]
\centering
\caption{Checks of the planted theorems and answer generators.}
\label{tab:verify}
\scriptsize
\begin{tabularx}{\textwidth}{@{}l X@{}}
\toprule
\textbf{Family} & \textbf{Check and result} \\
\midrule
F1 & Global balance holds symbolically ($K=2,3$; $L=3,4$). An exact solver equals the theorem on 60/60 random planted instances ($K$ 2--4, $L$ 3--8; largest difference $5.6\times10^{-16}$); all 60 violate detailed balance. Traps: the theorem's condition fails on 30/30, and every trap's law is at least $1.4\times10^{-3}$ from any product law. Monte Carlo (2M events) agrees within its interval. \\
F2 & The exact stationary law equals the transfer-matrix law on 48/48 random instances (largest relative difference $1.9\times10^{-14}$); the chains are irreducible. Traps are rejected 24/24, and the misapplied law is off by a median of 97\%. \\
F3 & Exact dynamic programming in rational arithmetic equals the planted penalty law on 48/48 checks (24 random instances, each for the number of steps and the penalty; $K\in\{2,3\}$, 4--8 particles). \\
(a) & Every answer is computed twice, from Conway's overlap numbers and from the pattern automaton's linear system, and the two must agree. \\
(b) & Every Burnside/P\'olya count is checked by orbit enumeration whenever the colouring space has at most $3\times10^5$ elements. \\
\bottomrule
\end{tabularx}
\end{table}

\section{Prompts and induction protocol}
\label{app:prompts}
\textbf{Solver.} Every condition uses the same prompt, ``Background notes:'' followed by the length-matched context, then ``Problem:'' and the statement, ending with ``Put your final answer in \texttt{\textbackslash boxed\{\}}.'' Contexts are cut or padded with neutral text to the token length of the oracle text, measured with the solver's tokenizer; a run is VOID if any context differs by more than 1\%.

\textbf{Contamination probe.} Before seeing any problem, each inducer receives the process described in words and two prompts, four samples each: \emph{recall} (``Do you know this process, or any published result about its long-run (stationary) distribution? If you know a closed form, or a condition on the rates under which it takes a simple form, state it precisely and say where it comes from \ldots'') and \emph{conjecture} (``Without running any code, conjecture what the stationary distribution of this process looks like, and state the condition on the rates under which your conjecture holds.''). For F3 and P5b the prompts name the family's own quantity (the expected penalty until the box is empty; the waiting times).

\textbf{Induction, planted families (protocol v2).} The inducer sees the process, one training statement in full, a listing of all training problems with their exact answers, and a working directory with every statement. It may run Python (numpy and the standard library) by writing one code block per message, and sees the output. Its goal, quoted: ``Your goal is NOT to answer these problems one by one. It is to discover the general structure behind the answers, in terms of the rates, and write a THEORY NOTE that lets a strong student solve NEW problems of this kind \ldots{} in a few lines without a computer, and that says when it does NOT apply. A THEORY NOTE is accepted only after you have run code at least 3 times.'' A note submitted earlier is refused with a reminder to test ideas against the exact answers first; after 24 messages, or when the token budget runs low, the inducer is told to write the note now. The note is at most two pages, with definitions, statements, scope and how to compute, and no answers to specific problems. Protocol v1 was identical except for the code gate and the inline statement: under it, the first weak run wrote all three notes on its first turn without reading the data.

\textbf{Induction, family (a) (searched).} The inducer writes 8 candidate theories from 30 ``fit'' training problems with answers (four prompt framings, two samples each); each candidate is placed in the solver's context and scored on the 10 remaining ``validation'' training problems ($k=4$); the best candidate is the induced theory, and a refined version is rewritten once from the solver's validation failures. A first variant also asks for an executable rule that reproduces the 40 training answers.

\textbf{Merging.} The merger sees the 12 induced notes and the 40 training problems with answers: ``Some notes contain correct rules, some contain wrong or incomplete ones, and different notes cover different problem types. Merge them into ONE theory note: keep every rule that agrees with the solved problems, fix or drop rules that do not, and make sure every problem type is covered. Check rules against the solved problems yourself.'' The merged note has the oracle text's word limit and may not copy any problem's numbers. Each merger is sampled three times.

\section{Judge protocol}
\label{app:judge}
The judge sees the planted theorem and one induced theory, with model names removed, and assigns one of five classes: (i) the same structure, (ii) equivalent but differently formulated, (iii) a partial or special case, (iv) wrong but a useful heuristic, (v) wrong. Each theory is judged three times and the majority class is reported; ``split'' means no majority. The preregistered primary judge is gpt-oss-120b~\cite{openai2025gptoss}, used for all four planted and classic families. For F2, F3 and P5b we also ran a secondary judge, Muse Spark 1.3, on identical anonymized inputs. Table~\ref{tab:judge} reports their agreement on the 21 induced theories of these families (three weak, three strong and, for F2, three theories of the original Qwen3-235B-A22B). Agreement is low overall and differs by family: on F3 both judges follow solver accuracy, on P5b only the secondary judge does, and on F2 every theory is at zero accuracy. The coarse classes merge (i) and (ii) into recovered, keep (iii) as partial, and merge (iv) and (v) into wrong.

\begin{table}[h]
\centering
\caption{Agreement between the primary (gpt-oss-120b) and secondary (Muse Spark 1.3) judges, and Spearman correlation of each judge's class with solver accuracy (classes ordered from wrong to recovered).}
\label{tab:judge}
\small
\begin{tabular}{@{}lcccccc@{}}
\toprule
\textbf{Family} & \textbf{Exact} & \textbf{Coarse} & $\kappa$ \textbf{exact} & $\kappa$ \textbf{coarse} & $\rho$ \textbf{primary} & $\rho$ \textbf{secondary} \\
\midrule
F2 & 4/9 & 4/9 & $-0.22$ & $-0.22$ & $0.21$ & $-0.41$ \\
F3 & 3/6 & 4/6 & $0.38$ & $0.56$ & $0.95$ & $0.93$ \\
P5b & 3/6 & 3/6 & $-0.29$ & $-0.29$ & $-0.62$ & $0.66$ \\
All & 10/21 & 11/21 & $0.11$ & $0.18$ & & \\
\bottomrule
\end{tabular}
\end{table}

\section{Compute and reproducibility}
\label{app:repro}
The follow-up experiments (the 32k and 64k re-evaluations, family (b), the stronger inducers, F2, F3, P5b and merging) used 128.2 node-hours on single GB200 and GB300 nodes, under a preregistered budget of 150 node-hours with at most four nodes at a time; preempted and voided jobs are included. The strong-inducer runs with Qwen3-235B-A22B-Thinking-2507 (FP8) ran on separate single-node H100 machines, and the judges on those machines or through an API; neither is included. The first family (a) and F1 experiments used additional single-node jobs, not included in this total.

\textbf{Models and sampling.} Solvers: Qwen3-14B and Qwen3.6-27B in thinking mode, and Qwen3-4B for family (b). Inducers: Qwen3-32B, the original Qwen3-235B-A22B (bf16, tensor parallel 4) and Qwen3-235B-A22B-Thinking-2507. Qwen models sample with temperature 0.6, top-$p$ 0.95 and top-$k$ 20; mergers use temperature 0.7; gpt-oss-120b uses temperature 1.0 and top-$p$ 1.0. Generation uses a fixed seed (0 for evaluation), and bootstrap intervals use 10{,}000 paired resamples. Table~\ref{tab:main} is generated from the grade and judge files by \texttt{build\_paper\_table.py}, and the figures by \texttt{figures/extract\_data.py} and \texttt{figures/make\_figures.py}; every number in the paper traces to a result file.

\end{document}